%% file: acl_latex.tex
\pdfoutput=1

\documentclass[11pt]{article}

\usepackage[]{acl}

\input{math_commands.tex}

\usepackage{times}
\usepackage{latexsym}
\usepackage{graphicx}
\usepackage{multirow}
\usepackage{color}
\usepackage{booktabs}
\usepackage[T1]{fontenc}

\usepackage[utf8]{inputenc}

\usepackage{microtype}
\usepackage{pifont}

\newcommand\our{PreQuant}

\title{PreQuant: A Task-agnostic Quantization Approach for \\ Pre-trained Language Models}

\author{ \textbf{Zhuocheng Gong\textsuperscript{1}\footnotemark[1], Jiahao Liu\textsuperscript{2}, Qifan Wang\textsuperscript{3}, Yang Yang\textsuperscript{2}, Jingang Wang\textsuperscript{2}, Wei Wu\textsuperscript{2} }\\
\textbf{Yunsen Xian\textsuperscript{2}, Dongyan Zhao\textsuperscript{1,4,5}\footnotemark[2], Rui Yan\textsuperscript{6,7}}\footnotemark[2] \\
\textsuperscript{1}Wangxuan Institute of Computer Technology, Peking University \\
\textsuperscript{2}Meituan; \textsuperscript{3}Meta AI \\
\textsuperscript{4}National Key Laboratory of General Artificial Intelligence \\
\textsuperscript{5}Beijing Institute for General Artificial Intelligence \\
\textsuperscript{6}Gaoling School of Artificial Intelligence, Renmin University of China \\
\textsuperscript{7}Engineering Research Center of\\ 
Next-Generation Intelligent Search and Recommendation, Ministry of Education \\
\texttt{\{gzhch,zhaody\}@pku.edu.cn}, \texttt{ruiyan@ruc.edu.cn}, \texttt{wqfcr@fb.com} \\
\texttt{\{liujiahao12,yangyang113,wangjingang02,xianyunsen\}@meituan.com} \\ \texttt{wuwei19850318@gmail.com}}

\begin{document}
\maketitle

\begin{abstract}
While transformer-based pre-trained language models (PLMs) have dominated a number of NLP applications, these models are heavy to deploy and expensive to use.
Therefore, effectively compressing large-scale PLMs becomes an increasingly important problem. 
Quantization, which represents high-precision tensors with low-bit fix-point format, is a viable solution. 
However, most existing quantization methods are task-specific, requiring customized training and quantization with a large number of trainable parameters on each individual task. 
Inspired by the observation that the over-parameterization nature of PLMs makes it possible to freeze most of the parameters during the fine-tuning stage, in this work, we propose a novel ``quantize before fine-tuning'' framework, \our, that differs from both quantization-aware training and post-training quantization. \our\ is compatible with various quantization strategies, with outlier-aware parameter-efficient fine-tuning incorporated to correct the induced quantization error. We demonstrate the effectiveness of \our\ on the GLUE benchmark using BERT, RoBERTa, and T5. We also provide an empirical investigation into the workflow of \our, which sheds light on its efficacy.
\end{abstract}

\renewcommand{\thefootnote}{\fnsymbol{footnote}}
\footnotetext[1]{Work done during internship at Meituan.}
\footnotetext[2]{Corresponding authors: Dongyan Zhao (zhaody@pku.edu.cn) and Rui Yan (ruiyan@ruc.edu.cn).}


\section{Introduction}

Pre-trained language models (PLMs) have shown superior performance in various NLP applications.
Despite their impressive success, these transformer-based models typically contain hundreds of millions of parameters. 
Massive model scale is becoming an increasing burden, preventing researchers from making full use of large-scale PLMs. 
According to a recent study, only $0.5$\% to $4$\% of research papers published at the recent five NLP conferences tend to adopt large PLMs (PLMs with over a billion parameters)~\citep{ding2022delta}. 
This suggests that the inefficiency of deploying large PLMs is hampering the development of NLP research.
Therefore, compressing PLMs becomes an urgent and important problem.

Various model compression methods have been proposed, such as knowledge distillation~\citep{jiao2020tinybert,sanh2019distilbert,wang2021minilmv2,passban2021alp}, weight sharing~\citep{lan2019albert}, network pruning~\citep{liang2021super,gordon2020compressing,li2021differentiable}, and quantization~\citep{tao2022compression,zhang2020ternarybert,bai2021binarybert,kim2021bert}.
Among these compression methods, quantization is a promising solution. The core idea of quantization is to use low bit precision to store weight and activation tensors, and use fixed-point arithmetic to speed up inference. There are some prior works on quantizing PLMs covering different strategies and granularities. However, these quantization methods generally neglect the characteristics of PLMs - the distinction between fine-tuning a model and training a model from scratch - but treat quantizing PLMs no different from quantizing regular neural networks. In other words, these methods are task-specific, which design customized quantization for PLMs. There are two main limitations: first, these task-specific methods need to conduct both quantization and fine-tuning for each downstream task, with the quantization being applied either during or after the fine-tuning stage, which is inefficient; Second, the number of trainable parameters are still very large during fine-tuning, which is computational expensive.

In this work, we consider the quantization pipeline specially for the pre-training scenario. Our motivation starts from the distinction between ``training from scratch'' and ``pre-training then fine-tuning''.
Unlike the weights from random initialization, the weights of the pre-trained model already contain rich information during pre-training. 
To utilize such information in a more efficient manner, we propose to directly quantize the pre-trained model in a task-agnostic way to obtain a ``pre-quantized'' model before fine-tuning.
We then introduce a parameter-efficient fine-tuning and show that fine-tuning could be finished with minimal weight updates. In particular, we freeze most of the quantized weights in the ``pre-quantized'' model, and only fine-tune a very small subset of its model parameters in the fine-tuning process. 
Through an extensive set of explorations and experiments, we demonstrate the feasibility and advantages of the ``quantizing the PLM first, then fine-tuning'' pipeline, which we name as \our. The main contributions are summarized as follows: 


\begin{itemize}
    \item We propose a novel quantization framework, \our, tailored for PLMs. We conduct a systematic study to overcome the difficulties of PLM quantization and validate the performance through thorough experiments. 
    \item \our\ performs task-agnostic quantization, which dramatically reduces the storage requirements for large PLMs and enables efficient deployment of PLMs on different downstream tasks. 
    Moreover, \our\ only fine-tunes $0.5$\% of the model parameters, which is more suitable in resource-limited scenarios.
    \item \our\ is highly flexible, which is compatible with a wide range of quantization strategies and fine-tuning techniques. Within this framework, we evaluate the pros and cons of various quantization strategies and discuss the impact of different quantization settings.
\end{itemize}

\section{Related Work}

\subsection{Efficient Transformer}
Compressing transformer-based models has been a prosperous topic since PLMs showed remarkable performance in various NLP tasks~\citep{ganesh2021compressing}. 
The main idea of model compression is to reduce the memory and computation consumptions without too much performance degradation.
There are several strands of research for large-scale transformers compression, including knowledge distillation~\citep{jiao2020tinybert,sanh2019distilbert,wang2021minilmv2,passban2021alp}, quantization~\citep{tao2022compression,zhang2020ternarybert,bai2021binarybert,kim2021bert}, weight sharing~\citep{lan2019albert} and network pruning~\citep{liang2021super,gordon2020compressing,li2021differentiable}. 
Besides directly compressing transformers, parameter efficient fine-tuning becomes promising by restricting the number of trainable parameters during fine-tuning~\citep{houlsby2019parameter,zaken2022bitfit,hu2021lora,gong2022finding}.
\our{} propose an outlier-aware parameter-efficient fine-tuning method in its second stage.
\begin{figure*}[t]
\begin{center}
\includegraphics[scale=0.38]{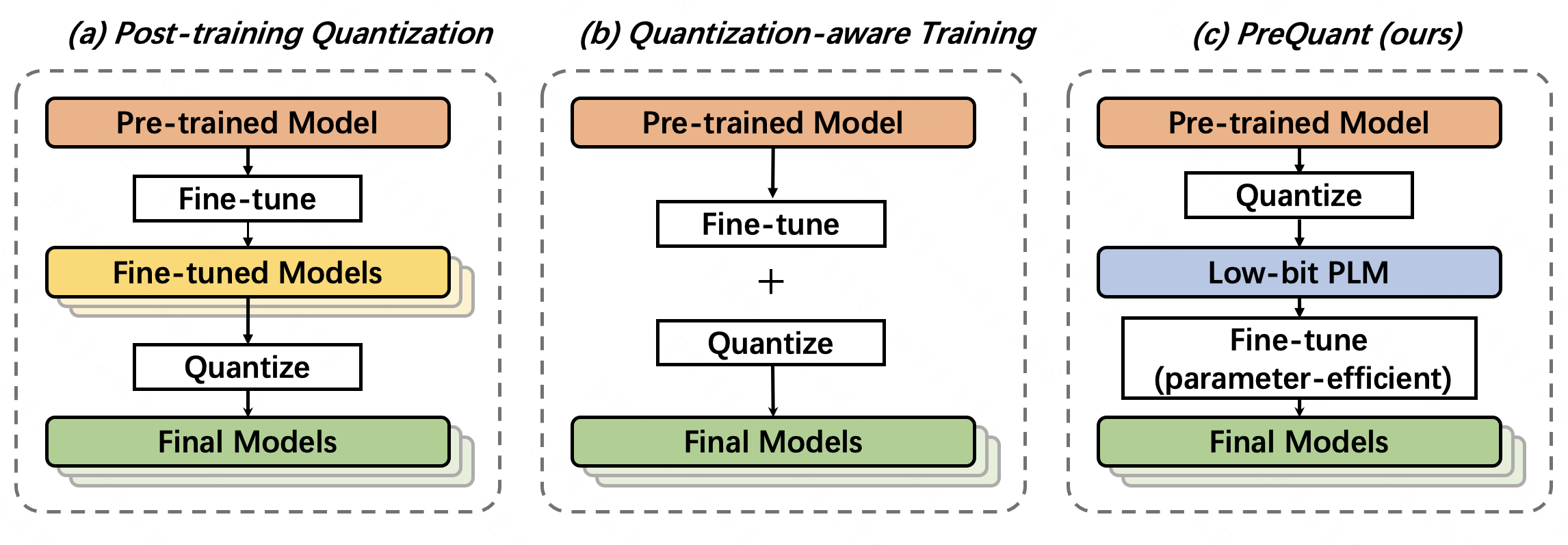}
\end{center}
\caption{An illustrative comparison of different quantization methods for PLMs. PTQ directly performs model qunatization after the fine-tuning stage, while QAT jointly optimizes qunatization and fine-tuning. In contrast, our \our\ conducts task-agnostic quantization first, and then performs parameter efficient fine-tuning.}
\label{fig:0}
\vspace{-4mm}
\end{figure*}
\subsection{Quantization}
Quantization, which represents the weights and activations of neural networks with low-bit precision, has been widely studied in computer vision and natural language processing (NLP) communities~\citep{gholami2021survey}.
Recently, some researchers attempt to compress PLMs to reduce the deployment cost with quantization methods~\citep{zadeh2020gobo,wu2022extreme,kim2021bert,bondarenko2021understanding}. 
Quantization-aware training (QAT)~\citep{gupta2015deep} is a representative approach to quantize a PLM while retaining most of its performance on downstream tasks. 
Given a downstream task, QAT performs the quantization during the task-specific training(i.e., fine-tuning) process.
For example, Q8BERT~\citep{zafrir2019q8bert} and Q-BERT~\citep{shen2020q} are typical QAT methods to compress BERT-based models. 
Unlike QAT, Post-training quantization (PTQ) disentangles the fine-tuning and quantization.
The quantization procedure is conducted after the task-specific fine-tuning. 
In comparison to QAT, PTQ holds the advantages of flexibility and good compatibility.
\citet{yao2022zeroquant} combines PTQ with knowledge distillation to achieve efficient compression for large PLMs.
In addition to NLP scenarios, PTQ is also utilized to compress vision transformers~\citep{liu2021post}.
Some very recent researches employ quantization and parameter-efficient fine-tuning jointly.
Qadapter~\citep{park2022quadapter} introduces a lightweight module to produce quantization-friendly activations by scaling them channel-wise. 
AlphaTuning~\citep{kwon2022alphatuning} utilizes binary-coding-quantization (BCQ) by only updating scaling factors.

\subsection{Outlier Phenomenon and its Applications in Quantization}
Outlier phenomenon in PLMs has been observed in previous research. 
\citet{kovaleva2021bert} reveals that PLMs are surprisingly fragile to the removal of a very small number of features in the layer outputs. 
More specifically, in case of BERT-based PLMs, outlier values exist in LayerNorm, the disabling of which would disrupt both the Masked Language Modeling (MLM) loss and the downstream task performance.
The outliers are high-magnitude normalization parameters that show up consistently in the same dimensional positions.
Outlier phenomenon has some applications in quantization.
For example, \citet{park2018energy} proposes to use a low-precision format for the center values and a high-precision format for the outliers in PTQ.
\citet{zhao2019improving} proposes an outlier channel splitting (OCS) method that duplicates and halves the channels containing outlier value.
\citet{bondarenko2021understanding} shows that outlier values detected in the activation of PLMs affect the estimation of corresponding scaling factors, thus disturbs the effectiveness of quantization. 
Hence, outlier-aware quantization has been proposed to promise the quantization performance.
In \our{}, we take the outlier phenomenon into consideration in both stages, which are first detected and then treated separately in low-precision quantization.
During the fine-tuning stage, we cast the outliers back to high-precision representations and only update them.

\section{Preliminary}
A number of works have been employing various quantization techniques on the field of pre-trained language models. 
Existing quantization methods can be categorized into two prominent branches: quantization-aware training and post-training quantization.

\begin{figure*}[t]
\begin{center}
\includegraphics[scale=0.4]{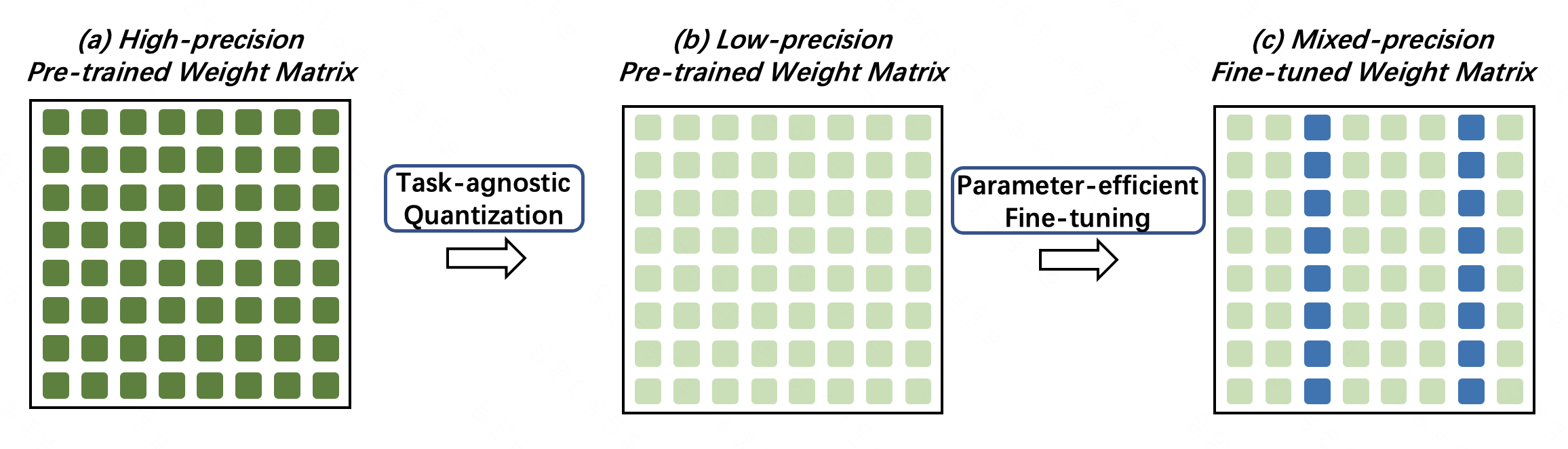}
\end{center}
\caption{The illustration of our two-stage quantization framework. 
Dark green and light green blocks represent for weight values in high-precision and low-precision respectively. Blue blocks represent for fine-tuned weights. In the first stage, all weights are ``pre-quantized'' to low-precision indiscriminately. 
In the second stage, a very small portion of weights are updated while the others are frozen during fine-tuning.}
\label{fig:1}
\end{figure*}

\paragraph{Basic Notations} 
We consider uniform quantization for both weights and activations. Specifically, for a given tensor $\textbf{x}$ in full precision, we adopt the rounding-to-nearest operation to round $\textbf{x}$ to the nearest unsigned integer grid values $\textbf{x}^{\sZ}$, which can be described as: 
\begin{equation}
\label{eq:1}
    \textbf{x}^{\sZ}= \text{clip}\left(\left\lfloor\frac{\textbf{x}}{\alpha}\cdot2^b\right\rceil+z;0,2^b-1\right)
\end{equation}
where $b\in\sN$ is bit-width, $\alpha\in\sR$ is the scaling factor, and $z\in\sN$ is zero-point. After obtaining the quantized tensor $\textbf{x}^{\sZ}$, one can approximate the full-precision version of the tensor $\widehat{\textbf{x}}$:
\begin{equation}
    \widehat{\textbf{x}}=\frac{\left(\textbf{x}^{\sZ}-z\right)\alpha}{2^b}
\end{equation}

\paragraph{Quantization-aware Training (QAT)}
QAT methods (\autoref{fig:0}(b)) learn the scaling factors (quantization) along with the weights during the fine-tuning stage. 
Since the rounding operation in \autoref{eq:1} is not derivable, gradients through the non-differentiable operations are usually approximated with the Straight-through Estimator (STE)~\citep{bengio2013estimating}.
As the quantization process of QAT is supervised by the overall training objective, the performance is generally quite promising.

\paragraph{Post-training Quantization (PTQ)}
PTQ methods (\autoref{fig:0}(a)) conduct qunatization after the fine-tuning.
Unlike QAT that relies on the full training data, PTQ requires very little sometimes even zero calibration data to estimate scaling factors. Therefore, the overhead of PTQ is relatively small. 
However, its ease of use often comes with significant performance penalties. 

Generally, existing quantization solutions (both QAT and PTQ) for PLMs are \textbf{task-specific}, meaning to quantize either during or after the model fine-tuning stage.
However, in PLMs, ``pre-training then fine-tuning'' replaces conventional ``training from scratch'', thus pre-trained weights already contain rich information. 
We wonder if it is possible to perform \textbf{task-agnostic} quantization. 
As shown in \autoref{fig:0}(c), \our\ first conducts task-agnostic quantization on the pre-trained model, followed by parameter-efficient fine-tuning.

\section{\our{}}

\subsection{Overview}
In contrast to PTQ and QAT, we propose to quantize PLMs prior to fine-tuning.
Specifically, our framework consists of two stages, as shown in \autoref{fig:1}. 
The first stage directly quantizes the pre-trained weights of PLMs without further adaptation. 
Hence, the quantization is task-agnostic.
The second stage fine-tunes the ``pre-quantized'' PLM for downstream tasks.
We can not simply apply the vanilla fine-tuning setting to a ``pre-quantized'' PLM.
When the vanilla fine-tuning setting is used, it converts low-precision weights back into high-precision representations as weight updates are necessarily in high-precision (low-precision training is practically impossible). This defeats our purpose of quantizing these values.
To address the issue, we propose a parameter-efficient tuning method that freezes most of the quantized weights and only fine-tune a small subset of model parameters.
The details would be presented in following sections.


\subsection{Task-agnostic Quantization}
\label{method:quant}
The goal of the uniform quantization in \autoref{eq:1} is to estimate the optimal scaling factor $\alpha$ for each parameter matrix. 
This can be formulated as an optimization problem that minimizes certain loss functions, such as mean squared error~\citep{choukroun2019low}. 
A more convenient solution is to directly estimate $\alpha$ with statistic information, such as directly utilizing the range of the tensor as the scaling factor~\citep{bondarenko2021understanding}.

In our investigation into the weights of PLMs, we have observed outlier phenomenon: in each parameter matrix of PLMs, a tiny fraction of weights (i.e., outliers) holds abnormally greater values than the other weights. 
Empirically, most of weights strictly follow Gaussian distribution while ``outliers'' falls into the tail of the distribution, which can be detected with:
\begin{equation} 
\small
    \mW_{outlier}=\left\{w\ \bigg|\ \frac{1}{\sqrt{2\pi\sigma^2}}e^{-\frac{(x-\mu)^2}{2\sigma^2}}>\epsilon,w\in \mW\right\},
\end{equation}
where $\mu$ and $\sigma^2$ are the mean and variance of the parameter matrix $\mW$.
Outlier values affect the effectiveness of quantization, causing great quantization error~\citep{kovaleva2021bert}.
This addresses this issue, we adopt an intuitive quantization method. 
We set the quantization scaling factor $\alpha$ to $6\sigma$, which is big enough to clip all the outlier weights according to \autoref{eq:1}.

It is worth noting that \our\ is compatible with the other methods. 
In addition to the aforementioned \textit{outlier-aware} scaling factor, we implement three other methods for comparison.
\begin{itemize}
    \item \textit{Min-max} is a basic method that estimates the scaling factor with the minimum and maximum of the tensor~\citep{vanhoucke2011improving}.
    \item \textit{MSE} optimizes the scaling factor by minimizing the mean squared error between quantized and full-precision tensors~\citep{choukroun2019low,shin2016fixed,zhao2019improving}.
    \item \textit{Row-wise quantization} adopts a finer granularity that assigns different scaling factors to each dimension of the matrix~\citep{shen2020q,bondarenko2021understanding}. 
\end{itemize}

We conduct a thorough comparison on previous scaling factor estimation methods and discuss the advantages and disadvantages of each in the experiment section.
In comparison to previous quantization methods, our quantization method is data-free and task agnostic, as the quantizations are executed directly prior to the fine-tuning.

\subsection{Outlier-aware Parameter-efficient Fine-tuning}
\label{method:ft}
After obtaining a ``pre-quantized'' PLM, the second stage is to fine-tune the model for specific downstream tasks.
In this stage, we encounter a dilemma: on one side, fine-tuning requires updating model weights with high-precision representations, while on the other side, casting the low-precision weights back to high-precision weights will nullify the effect of quantization.
To address the issue, we propose an outlier-aware parameter-efficient fine-tuning (outlier-aware tuning) strategy that keeps most of the model parameters frozen in low-precision.
Parameter-efficient fine-tuning aims to adapt PLMs by tuning only a few number of parameters~\citep{houlsby2019parameter,gong2022finding}. 
\citep{zaken2022bitfit} and \citet{gong2022finding} have shown that tuning a small subset of parameters of PLMs can be comparable with full-parameter fine-tuning in terms of performance. This approach suits our scenario as it does not modify the model structure.

However, parameter-efficient fine-tuning is more challenging in our case since the quantization step produces pre-quantized PLM wherein the weights are rounded to low-precision. 
The induced quantization error correlates to the disturbance of the weights. If weights do not change much after quantization, the error will be minimal, and significant if they do. 
Intuitively, our goal is to identify which parts of the weights cause the most quantization error. By only tuning these specific weights, we can recover much of PLM's damaged representation ability.

In our investigation, we find that the majority of parameters exhibit relatively small disturbance, hence freezing them could preserve most of the PLM’s ability. 
Some particular weights contribute to the most of the induced error and these weights are concentrated in specific dimensions. Moreover, these susceptible-to-quantization weights are exactly outlier weights that we mentioned in the above section. This is because the abnormally large values of outliers are generally clipped according to \autoref{eq:1}.
We identify the dimensions containing most of outlier weights, then setting them as trainable parameters while freezing the rest. 
Specifically, in each parameter matrix, we select $r$ outlier dimensions as trainable parameters. 
$r$ is extremely small, we can guarantee that more than $99$\% parameters still remain in low-precision.
By tuning the subnetwork consisting of outlier dimensions, we expect to recover the damaged representation ability and adapt to specific downstream task at minimal trainable parameters.

\input{tables/main}
\input{tables/main2}

\section{Experimental Evaluation}
\subsection{Experimental Setup}
\paragraph{Settings} 
We evaluate \our\ on several popular PLMs including BERT~\citep{devlin2018bert}, RoBERTa~\citep{liu2019roberta} and T5~\citep{raffel2020exploring}. 
For RoBERTa, we test on both $\text{RoBERTa}_{\text{base}}$ and $\text{RoBERTa}_{\text{large}}$. 
For T5, we employ \our\ to the encoder of $\text{T5}_{\text{3b}}$, denoted as T5 Encoder. 
We use a fixed set of hyper-parameters for all the GLUE tasks. 
For each layer, we set the bit-width option $b$ for weights as $4$. 
Besides, we apply $8$-bit min-max uniform quantization to activations and embeddings. 
Experimental results of more bit-width options are listed in Appendix~\ref{sec:bit}.

\paragraph{Datasets} 
The GLUE benchmark contains a variety of natural language understanding tasks, including textual entailment (RTE), natural language inference (MNLI, QNLI), paraphrase (MRPC, QQP, STS-B), sentiment analysis (SST-2) and linguistic acceptability (CoLA)~\citep{wang2018glue}. 
The evaluation metrics are Matthews correlation for CoLA, Spearman correlation for STS-B, and Accuracy for the other tasks. 
We supplement fine-tuning details in Appendix~\ref{sec:trainingdetail}.

\paragraph{Baselines}
Classical quantization methods including PTQ and QAT are adopted as baselines.
For PTQ, we adopt the implementation by~\citet{bondarenko2021understanding}, which introduces the group-wise granularity to reduce the quantization error. 
For QAT, we also implement a group-wise granularity variant. 
Results of the vanilla QAT that utilizes straight-through estimator (STE)~\citep{bengio2013estimating} to spread gradients are listed in Apppendix~\ref{sec:bit}.
We include Qadapter~\citep{park2022quadapter} and AlphaTuning~\citep{kwon2022alphatuning} that jointly employ the quantization and the parameter-efficient fine-tuning for further comparison.

\subsection{Main Results}
\paragraph{Comparison with Quantization Methods.}
The main comparison results are reported in \autoref{tab:main}.
Due to the precision reduction, all quantization methods inevitably lead to performance degradation in comparison to the full-precision fine-tuned model (FT).
There is a considerable performance gap between $4$-bit PTQ and $32$-bit FT, although they are both tuning with a modest amount of calibration data.
QAT outperforms PTQ on all tasks, demonstrating the benefit of a hybrid approach of quantization and task-specific fine-tuning.
\our{} is comparable in performance to QAT, but with much fewer trainable parameters.
In order to evaluate the scalability and robustness of \our, we report the results for different scale PLMs, ranging from $110$M parameters to $1.5$B parameters.
As the model size increases, \our{} performs consistently and stably.
Take $\text{T5}_{\text{1.5b}}$ as an example, \our{} could achieve $99.21$\% performance of FT with only tuning $0.10$\% trainable parameters.

\paragraph{Comparison with Parameter-efficient PLM Quantization Methods.}
Comparisons with Qadapter and AlphaTuning are reported in \autoref{tab:main2}.
For Qadapter, we adopt uniform asymmetric $8$-bit channel-wise quantization for both activation functions and weights as described in the original paper. 
We implement AlphaTuning with $4$-bit BCQ quantization to make a fair comparison. 
Overall, \our{} achieves the best performance among these parameter-efficient PLM quantization methods, while maintaining a comparable compression ratio.
Inspired by AlphaTuning, we also implement \our-$\alpha$, a variant of \our\ that only tuning the scaling factors of the uniform quantization, to estimate the value of AlphaTuning technique. 
\our\ outperforms \our-$\alpha$ by $1$ point, indicating the advantage of updating the model parameters over updating the quantization parameters.


\begin{figure}[t]
\begin{center}
\includegraphics[scale=0.35]{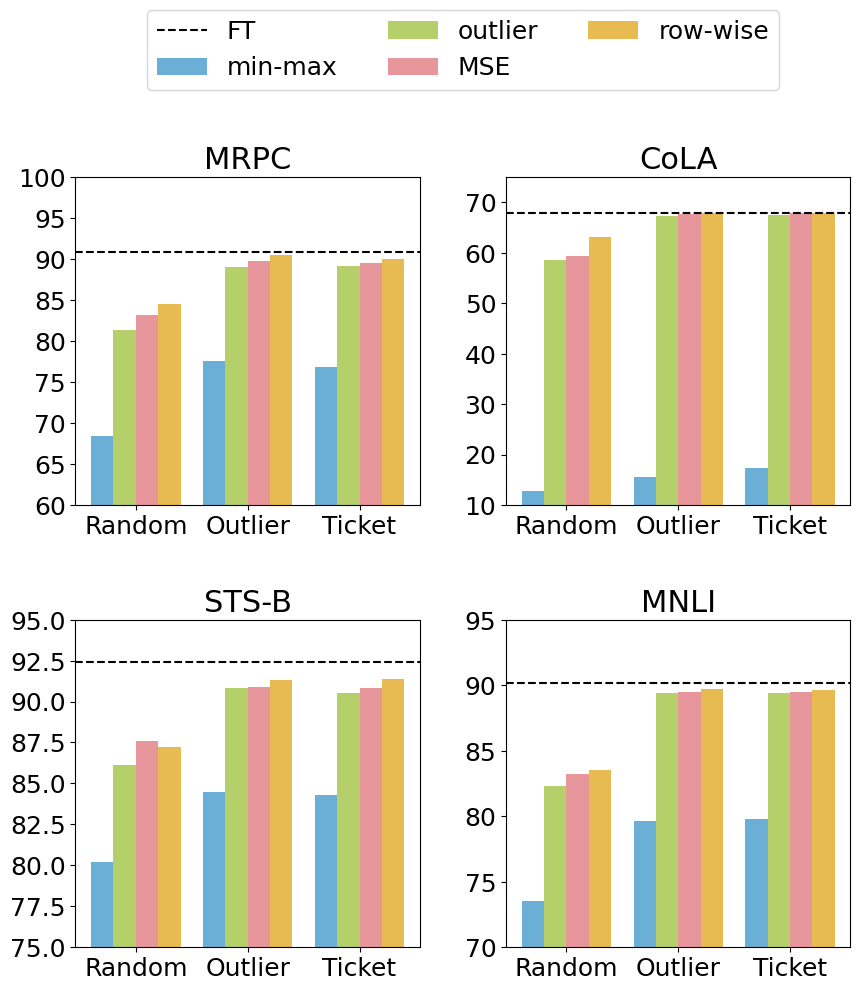}
\end{center}
\caption{An comparison of different quantization strategies for PLMs. The pre-trained model is $\textbf{RoBERTa}_{\text{large}}$ and the bit-width is $4$.}
\label{fig:ab}
\end{figure}

\subsection{Comparison of Quantization Strategies}
\input{tables/analysis}

In this section, we replace the \textit{outlier-aware quantization} with alternative quantization strategies to see how different strategies affect the performance. 
We evaluate three different strategies (i.e., \textit{min-max}, \textit{MSE}, and \textit{Row-wise quantization} in Section~\ref{method:quant}) on $4$-bit quantization for $\text{RoBERTa}_{\text{large}}$. 
The differences of these strategies are listed in \autoref{tab:ab}. 
As the disturbance of weights after quantization indicates the induced quantization error, we compute the L2 distance between quantized weights and full-precision weights as the measurement of the quantization error. 
As the bottom block of \autoref{tab:ab} reveals, the induced quantization error is highly correlated to the performance on downstream tasks.
The less the error, the better the performance.
The \textit{min-max quantization} strategy performs worst due to the negative influence of outlier weights.
Meanwhile, \textit{outlier-aware}, \textit{MSE}, and \textit{row-wise} strategies achieve comparable performance on four tasks as well as similar quantization error. 
The \textit{MSE quantization} strategy achieve slightly better performance since it directly optimizes the L2 distance, which is more complicated than statistical strategies.
\textit{row-wise quantization} perform slightly better than layer-wise strategies at the cost of a more expensive computational graph.
Above all, the \textit{outlier-aware} strategy reaches the best trade-off between performance and complexity.

\subsection{Analysis of Outlier-aware Fine-tuning}
\label{sec:peft}
In this section, we discuss the effect of parameter-efficient fine-tuning on \our.

\paragraph{Does outlier-aware tuning really work?\ }
\our\ appoints the trainable subnetwork by detecting outlier dimensions, shorted as \textit{Outlier}. 
It is important to show that the outlier dimension really matters for fine-tuning performance. 
To this end, we introduce two variants: 1) \textit{Random}: We randomly choose the same amount of trainable parameters as our method; 2) \textit{Ticket}: This is a task-agnostic subnetwork for parameter-efficient fine-tuning proposed in \citet{gong2022finding}. 
The experimental results on four datasets are shown in \autoref{fig:ab}. 
Random selection of trainable parameters leads to a significant drop in performance, suggesting that outlier information does help in finding suitable trainable subnetworks.
\textit{Outlier} and \textit{Ticket} achieve comparable performance, and both are very close to the upper-bound performance by the FT.
This suggests that our \textit{outlier-aware} fine-tuning is a promising strategy to efficiently adapt PLMs to downstream tasks while reducing quantization errors. 
Noting that \textit{Outlier} and \textit{Ticket} have similar performance, we further calculate the subnetwork overlap ratio of the two methods using the Jaccard similarity coefficient.
As we expected, \textit{Outlier} and \textit{Ticket} have non-negligible overlap (Jaccard similarity coefficient is $0.57$.).

\input{tables/size}

\paragraph{What is the optimal size of the trainable subnetwork?\ }
As stated in Section~\ref{method:ft}, we use hyper-parameter $r$ to control the size of the trainable high-precision parameters. 
We then focus on the effect of $r$ on model performance. 
We conduct empirical experiments with various values of $r$ in \{$1, 3, 5, 10, 20, 512, 1024$\}. 
Smaller value of $r$ indicates fewer trainable parameters, which inevitably leads to performance degradation.
We expect that more trainable parameters will lead to higher performance.
The results are reported in \autoref{tab:size}. 
We find that a relatively small $r$, e.g., $3$ or $5$, is good enough to adapt \our\ to downstream tasks. 
Note that $r=512$ sets half of the model parameters trainable, and $r=1024$ denotes that the whole model is trainable. 
From \autoref{tab:size}, we can see that setting $r$ as $1024$ cannot fully recovers the performance which is reasonable because the induced quantization error between high-precision and low-precision representations could not be completely eliminated.
Setting $r$ to a larger value than $10$ brings limited performance improvements but requiring more high-precision computational cost.

\paragraph{Does other parameter-efficient fine-tuning methods work with \our\ ?}
Following~\citet{ding2022delta}, we consider three types of parameter-efficient techniques: addition-based methods, specification-based methods, and reparameterization-based methods. Addition-based methods, such as adapter and prefix-tuning, involve introducing extra trainable modules or parameters that cannot be directly applied to \our. On the other hand, specification-based methods specify certain parameters in the original model as trainable parameters, which work well with \our as discussed in Figure~\ref{fig:ab}. Our outlier-aware fine-tuning falls into this category. Reparameterization-based methods, such as low-rank adaptation (LoRA)~\citep{hu2021lora}, reparameterizes linear layers. LoRA updates all parameters in the weight matrix by adding a low-rank matrix. In our scenario, the original weight matrix is in low-precision while the update matrix is in high-precision. The addition of a high-precision matrix to a low-precision matrix results in a high-precision matrix, thus nullifying the quantization effect.

\subsection{Extending to Layer-wise Mixed-precision Quantization}
Previous work has shown that allocating different bit-widths to different layers leads to a better accuracy-efficiency trade-off, since not all layers are equally sensitive to quantization~\citep{tang2022mixed}.
\our\ can be conveniently extended to a layer-wise mix-precision variant by assigning customized bit-widths to each transformer layer. 
We implement a pilot mix-precision quantization paradigm that assigns $2$-bits to bottom layers and $4$-bits to top layers or vise versa. 
As can be seen in \autoref{tab:mp}, all mixed-precision methods exhibit performance degradation due to the hybrid quantization setting. 
An overall conclusion is that top layers are less sensitive to quantization than bottom layers. 
Allocating $2$-bits to the top third of layers resulted in an average loss of less than $3$ points, which is very impressive.
Meanwhile, assigning $2$-bits to the bottom one-third of the layers suffers from more than $10$ points of performance loss.
These insightful findings could be beneficial to the development of better mixed-precision quantization techniques.
\input{tables/mp}

\section{Conclusions}
As the scale of pre-trained language models increases, model compression becomes a prerequisite prior to model deployment in resource-limited scenarios.
Quantization is an effective and promising technique to compress large PLMs.
Existing quantization methods including PTQ and QAT perform quantizations either during or after task-specific fine-tuning process.
Since these approaches are highly task-specific, it's hard to transfer them to different tasks with low cost. 
In this paper, we propose a ``quantizing the PLM first, then fine-tuning'' framework, \our, which includes a task-agnostic quantization stage and an outlier-aware parameter-efficient fine-tuning stage.
We compress widely used PLMs with \our{}, including BERT, RoBERTa and T5 variants.
The experimental results on the GLUE benchmark are reported to demonstrate the effectiveness of \our{}.
We also reveal that \our{} is more flexible and efficient than its competitive counterparts.
An elaborate empirical study is conducted on the workflow of \our{}, we hope the findings could shed some light on the quantization research of PLMs.


\section*{Limitations}
Although the proposed \our\ achieves promising results especially in reducing the storage and computational resources, we discuss some limitations of our work in this section. 
In our experiments, we observe that the performance of \our\ is highly correlated with the data size. 
When fine-tuning with very limited data, \our\ may not meet expectation to preserve the performance of PLMs. 
Moreover, our model performance also depends on the number of parameters (i.e. outliers) restored in the fine-tuning stage. 
This hyper-parameter controls the trade-off between model performance and parameter efficiency. 
The optimal choice of the hyper-parameter for different tasks requires further investigation.
Additional discussion and experimental results are provided in Appendix~\ref{low_resource}. 


\section*{Acknowledgments}
This work is supported by Ministry of Science and Technology Key R\&D Program (2030 Artificial Intelligence) (No. 2020AAA0106600) and National Natural Science Foundation of China (NSFC Grant No. 62122089). We sincerely thank all reviewers for their valuable comments and suggestions, which are crucial for improving our work.
\bibliography{anthology,custom}
\bibliographystyle{acl_natbib}

\clearpage
\appendix
\section{Appendix}

 \begin{figure}[h]
\begin{center}
\includegraphics[scale=0.3]{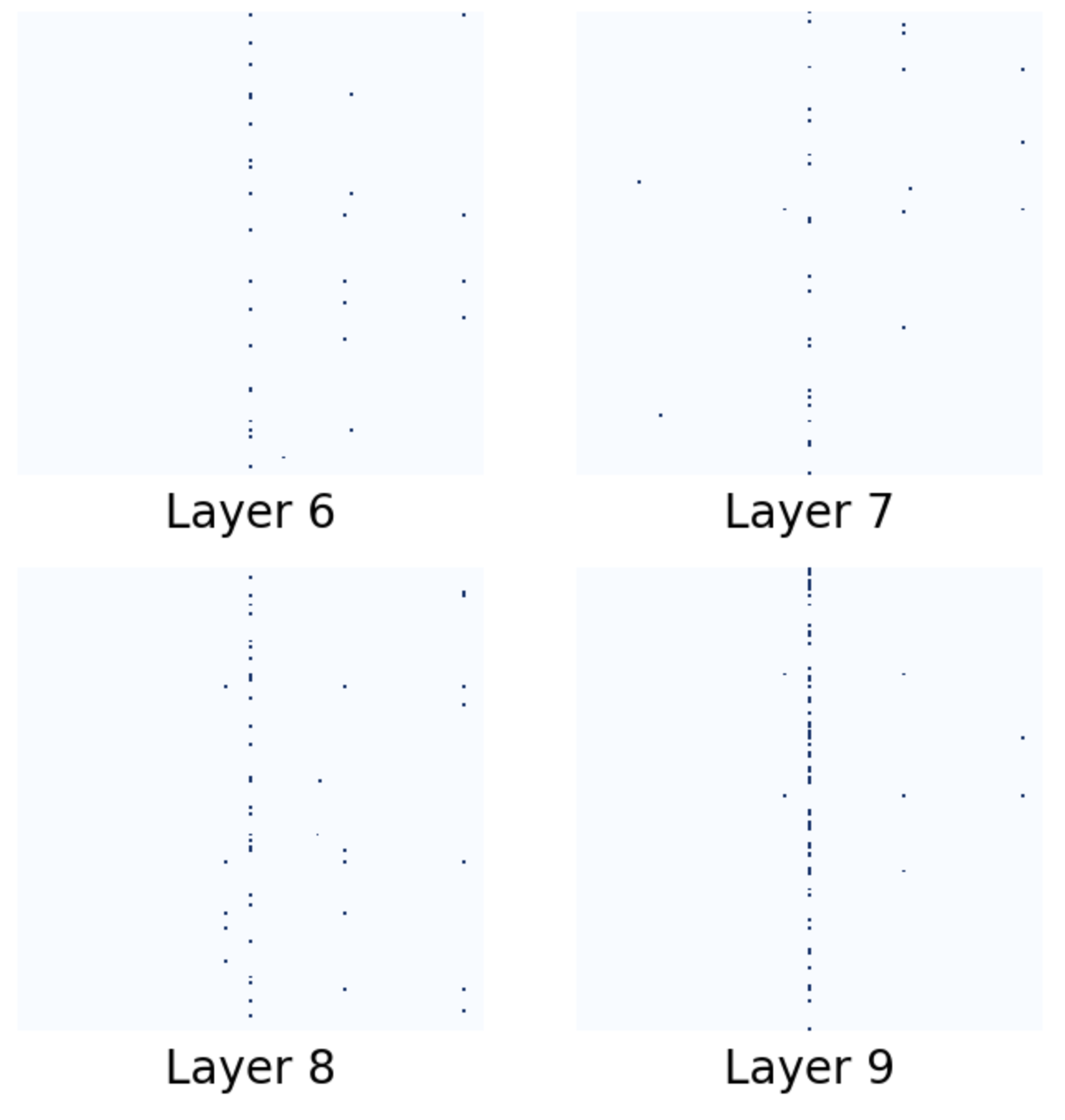}
\end{center}
\caption{Visualization of quantization error on $\textbf{BERT}_{base}$ key matrix from several layers. We compare the weights before and after the quantization, and plot the positions with large difference.}
\label{fig:noise}
\end{figure}

\subsection{Training Details}
\label{sec:trainingdetail}
For all the tasks, we adopt AdamW~\citep{loshchilov2018decoupled} as the optimizer and search batch size in \{16, 32\}. For full-parameter fine-tuning baselines, the learning rate for \our\ is searched within \{1e-5, 2e-5, 3e-5, 4e-5\} for $\text{BERT}_{\text{base}}$, $\text{RoBERTa}_{\text{base}}$, and $\text{RoBERTa}_{\text{large}}$ and \{1e-4, 2e-4, 3e-4\} for $\text{T5 Encoder}$. For \our, the learning rate is searched within \{1e-4, 3e-4, 5e-4, 7e-4, 9e-4\}. We set the dropout rate to 0.1 and weight decay to 0.01. For all tasks, the model is trained for 10 epochs at maximum and the best performance on the validation set is reported. Experiments are conducted upon the Huggingface Transformers library~\citep{wolf-etal-2020-transformers}.

\subsection{Results in Low-resource Scenarios}\label{low_resource}
\input{tables/data}

During investigation, we find quantization is more challenging in small datasets. We further explore the effect of data size on quantization and fine-tuning.
To this end, we randomly sample MNLI training set to \{2k, 4k, 6k, 8k\} examples and fine-tune \textbf{T5 Encoder} on them. As seen in \autoref{tab:data}, smaller data size leads to larger performance gap between the full-precision model and the quantized one.

\subsection{Visualization of Quantization Error}
\autoref{fig:noise} is an example of quantization error induced by uniform quantization. Several outlier dimensions tend to have larger error after quantization due to their large value.

\subsection{Scaling to Other Bit-widths}
\label{sec:bit}
\begin{figure}[h]
\begin{center}
\includegraphics[scale=0.45]{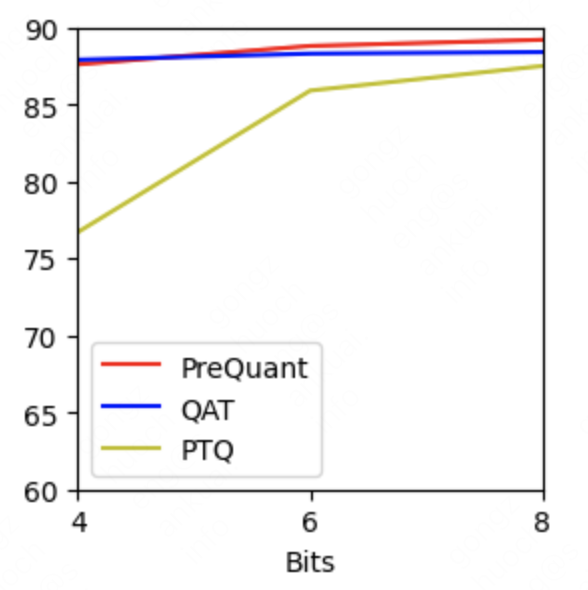}
\end{center}
\caption{Averaged performance of all GLUE tasks on $\textbf{RoBERTa}_{\text{large}}$.}
\label{fig:bit}
\end{figure}

As shown in  \autoref{fig:bit}, when the number of bits for weight is 8, the performance of all quantization methods is close. However, when the bit-width decreases to 4, performance disparities between various approaches start to become apparent. PTQ fails to predict reasonable answers on 4-bit quantization, indicating that the quantization error is too strong to be minimized with a modest amount of calibration data. QAT and \our\ still remain an acceptable performance for 4-bit quantization.

\subsection{Detailed Results of More Quantization Methods}
\label{sec:fullbaseline}
\input{tables/baseline}

\end{document}

%% file: math_commands.tex
\usepackage{amsmath,amsfonts,bm}

\def\eqref#1{equation~\ref{#1}}

\def\1{\bm{1}}

\def\mW{{\bm{W}}}

\DeclareMathAlphabet{\mathsfit}{\encodingdefault}{\sfdefault}{m}{sl}
\SetMathAlphabet{\mathsfit}{bold}{\encodingdefault}{\sfdefault}{bx}{n}

\def\sN{{\mathbb{N}}}

\def\sR{{\mathbb{R}}}

\def\sZ{{\mathbb{Z}}}



%% file: tables/main.tex
\begin{table*}[t]
\begin{center}
\setlength{\tabcolsep}{1.2mm}{
\scalebox{0.87}{\begin{tabular}{llrrrrrrrrrrr}
\toprule
\multirow{2}{*}{ Models} &\multirow{2}{*}{ Methods} & \multirow{2}{*}{ Bits}  &  Trainable &\multirow{2}{*}{ CoLA} &\multirow{2}{*}{ MNLI} &\multirow{2}{*}{ MRPC} &\multirow{2}{*}{ QNLI} &\multirow{2}{*}{ QQP} &\multirow{2}{*}{ RTE} &\multirow{2}{*}{ SST-2} &\multirow{2}{*}{ STS-B} &\multirow{2}{*}{ Avg.} \\
&&&  Params
\\ \midrule 
\multirow{4}{*}{$\text{BERT}_{\text{base}}$}& FT & $32$ & $85$M & $57.3$ & $84.4$ & $88.3$ & $91.6$ & $89.8$ & $71.0$ & $93.0$ & $89.4$ & $83.1$ \\
\cline{2-13}
& PTQ & $4$ & $85$M & $43.1$ & $68.2$ & $84.9$ & $79.7$ & $79.4$ & $50.2$ & $90.8$ & $83.1$ & $72.4$ \\
& QAT & $4$ & $85$M & $57.2$ & $83.7$ & $87.8$ & $91.3$ & $89.6$ & $70.0$ & $92.3$ & $89.1$ & $82.6$\\
& PreQuant & $4$ & \textbf{0.55M} & $54.6$ & $83.5$ & $88.0$ & $90.7$ & $88.6$ & $68.7$ & $92.3$ & $88.9$ & $81.9$\\
\hline
\multirow{4}{*}{$\text{RoBERTa}_{\text{base}}$}& FT & $32$ & $85$M & $63.6$ & $87.6$ & $90.2$ & $92.8$ & $91.9$ & $78.7$ & $94.8$ & $91.2$ & $86.4$ \\
\cline{2-13}
& PTQ & $4$ & $85$M & $46.3$ & $74.5$ & $85.5$ & $81.8$ & $84.3$ & $56.9$ & $92.1$ & $84.5$ & $75.7$ \\
& QAT & $4$ & $85$M & $61.9$ & $86.9$ & $88.9$ & $91.7$ & $91.3$ & $76.5$ & $94.4$ & $90.5$ & $85.3$ \\
& PreQuant & $4$ & \textbf{0.55M} & $61.5$ & $86.2$ & $89.0$ & $91.6$ & $90.9$ & $76.0$ & $94.0$ & $90.1$ & $84.9$\\
\hline
\multirow{4}{*}{$\text{RoBERTa}_{\text{large}}$}& FT & $32$ & $302$M & $68.0$ & $90.2$ & $90.9$ & $94.7$ & $92.2$ & $86.6$ & $96.4$ & $92.4$ & $88.9$ \\
\cline{2-13}
& PTQ & $4$ & $302$M & $46.6$ & $79.5$ & $86.6$ & $82.2$ & $84.6$ & $56.4$ & $92.6$ & $85.0$ & $76.7$ \\
& QAT & $4$ & $302$M & $66.5$ & $89.4$ & $88.8$ & $93.8$ & $91.4$ & $86.6$ & $95.8$ & $91.4$ & $87.9$\\
& PreQuant & $4$ & \textbf{1.47M} & $67.3$ & $89.4$ & $89.0$ & $93.2$ & $91.1$ & $84.7$ & $95.4$ & $90.8$ & $87.6$\\
\hline
\multirow{4}{*}{\text{T5 Encoder}}& FT & $32$ & $1.2$B & $67.6$ & $91.2$ & $90.9$ & $95.4$ & $91.9$ & $87.1$ & $97.2$ & $92.3$ & $89.2$ \\
\cline{2-13}
& PTQ & $4$ & $1.2$B & $50.6$ & $82.4$ & $86.5$ & $84.6$ & $85.7$ & $59.1$ & $92.0$ & $87.5$ & $78.6$\\
& QAT & $4$ & $1.2$B & $66.5$ & $90.4$ & $90.2$ & $95.3$ & $91.6$ & $86.6$ & $96.7$ & $91.6$ & $88.6$\\
& PreQuant & $4$ & \textbf{11.80M} & $66.4$ & $90.7$ & $90.0$ & $95.1$ & $92.0$ & $85.1$ & $96.9$ & $91.6$ & $88.5$ \\
\bottomrule
\end{tabular}}}
\end{center}
\caption{Results on the development set of the GLUE benchmark. 
We also report the quantity of trainable parameters (without embeddings) of each method.
FT represents for full-precision full-parameter fine-tuning, which achieves best performance as expected.
All the quantization methods are implemented with $4$-bits precision representations. }
\label{tab:main}
\vspace{-1mm}
\end{table*}

%% file: tables/main2.tex
\begin{table}[t]
\begin{center}
\setlength{\tabcolsep}{1.2mm}{
\scalebox{0.86}{\begin{tabular}{lrrr}
\toprule
Methods & Bits & Params & GLUE \\
\midrule
FT & $32$ & $302$M & $88.9$ \\
Qadapter~\citep{park2022quadapter} & $8$ & \textbf{0.29M} & $85.1$ \\
AlphaTuning~\citep{kwon2022alphatuning} & $4$ & $1.18$M & $86.3$\\
PreQuant-$\alpha$ & $4$ & \textbf{0.29M} & $86.6$\\
PreQuant & $4$ & $1.47$M & \bf $87.6$\\
\bottomrule
\end{tabular}}}
\end{center}
\caption{Results of parameter-efficient PLM quantization methods on $\text{RoBERTa}_{\text{large}}$. Full results are supplemented in Appendix~\ref{sec:fullbaseline}.}
\label{tab:main2}
\vspace{-1mm}
\end{table}


%% file: tables/analysis.tex
\begin{table}[t]
\begin{center}
\setlength{\tabcolsep}{1.2mm}{
\scalebox{0.88}{\begin{tabular}{lrrrr}
\toprule
\multirow{2}{*}{} & Min- & Outlier- & \multirow{2}{*}{MSE} & Row- \\
& max & aware &  & wise \\
\midrule
Layer-wise & \multirow{2}{*}{\color{red}\ding{52}} & \multirow{2}{*}{\color{red}\ding{52}} & \multirow{2}{*}{\color{red}\ding{52}} & \multirow{2}{*}{\color{blue}\ding{56}} \\
Granularity \\
\specialrule{0cm}{3pt}{2pt}
Statistical & \multirow{2}{*}{\color{red}\ding{52}} & \multirow{2}{*}{\color{red}\ding{52}} & \multirow{2}{*}{\color{blue}\ding{56}} & \multirow{2}{*}{\color{red}\ding{52}} \\
Strategy \\
\specialrule{0.25pt}{2pt}{2pt}
Quantization & \multirow{2}{*}{163.1} & \multirow{2}{*}{59.5} & \multirow{2}{*}{42.6} & \multirow{2}{*}{41.7} \\
Error (L2 Dist) \\
\specialrule{0pt}{2pt}{2pt}
CoLA & $15.6$ & $67.3$ & $67.8$ & $68.0$ \\
MRPC & $77.6$ & $89.0$ & $89.8$ & $90.0$ \\
STS-B & $84.5$ & $90.8$ & $90.9$ & $91.3$ \\
MNLI & $79.6$ & $89.4$ & $89.5$ & $89.7$ \\
\bottomrule
\end{tabular}}}
\end{center}
\caption{Comparison of different quantization strategies on $4$-bits \our\ of $\textbf{RoBERTa}_{\text{large}}$.}
\label{tab:ab}
\end{table}

%% file: tables/size.tex
\begin{table}[t]
\begin{center}
\scalebox{0.83}{\begin{tabular}{llrrr}
\toprule
\multirow{2}{*}{ Models} &\multirow{2}{*}{ Size} &  \# Trainable &\multirow{2}{*}{ QNLI} &\multirow{2}{*}{ MRPC}\\
&&  Ratio
\\ \midrule 
\multirow{6}{*}{$\text{RoBERTa}_{\text{large}}$}& FT & $100$\% & $94.7$ & $90.9$ \\
& r = $1024$ & $100$\% & $92.9$ & $90.3$ \\
& r = $512$ & $50$\% & $92.8$ & $90.2$\\
& r = $20$ & $1.95$\% & $93.6$ & $89.4$ \\
& r = $10$ & $0.98$\% & $93.3$ & $89.1$ \\
& r = $5$ & $0.49$\% & $93.2$ & $89.0$ \\
& r = $3$ & $0.29$\% & $93.2$ & $88.8$ \\
& r = $1$ & $0.10$\% & $86.5$ & $80.5$ \\
\midrule
\multirow{6}{*}{\text{T5 Encoder}}& FT & $100$\% & $95.4$ & $90.9$ \\
& r = $1024$ & $100$\% & $94.1$ & $90.4$ \\
& r = $512$ & $50$\% & $94.2$ & $90.4$ \\
& r = $20$ & $1.95$\% & $95.1$ & $90.2$ \\
& r = $10$ & $0.98$\% & $95.1$ & $90.0$ \\
& r = $5$ & $0.49$\% & $94.6$ & $88.9$ \\
& r = $3$ & $0.29$\% & $92.3$ & $86.7$ \\
& r = $1$ & $0.10$\% & $87.5$ & $79.4$ \\
\bottomrule
\end{tabular}}
\end{center}
\caption{Validation results on QNLI and MRPC after applying 4-bits \our\ to $\textbf{RoBERTa}_{\text{large}}$ and \textbf{T5 Encoder}. The FT line is the result of full-precision full-parameter fine-tuning.}
\label{tab:size}
\end{table}

%% file: tables/mp.tex

\begin{table}[t]
\begin{center}
\scalebox{0.8}{\begin{tabular}{lrrrrr}
\toprule
\multirow{2}{*}{ Methods} & \multicolumn{3}{c}{ Layers} & \multirow{2}{*}{ QNLI} & \multirow{2}{*}{ STS-B} \\
& $1$-$8$ & $9$-$16$ & $17$-$24$
\\ \midrule
FT & $32$ & $32$ & $32$ & $95.4$ & $92.3$\\
All 4-bits & $4$ & $4$ & $4$ & $95.1$ & $91.6$\\
Bottom one-third & $2$ & $4$ & $4$ & $84.9$ & $75.0$\\
Bottom two-thirds & $2$ & $2$ & $4$ & $82.4$ & $59.5$\\
Top one-third & $4$ & $4$ & $2$ & $92.3$ & $89.6$\\
Top two-thirds & $4$ & $2$ & $2$ & $84.7$ & $85.4$\\
\bottomrule
\end{tabular}}
\end{center}
\caption{Layer-wise mixed-precision quantization results for \textbf{T5 Encoder} on QNLI and STS-B. For the model with 24 layers, we quantize top (or bottom) one(or two)-third(s) layers to 2-bits while keeping the rest of the model in 4-bits.}
\label{tab:mp}
\end{table}

%% file: tables/data.tex
\begin{table}[h]
\begin{center}
\scalebox{0.9}{\begin{tabular}{llllll}
\toprule
 Dataset Size & $2$k &  $4$k &  $6$k &  $8$k &  full 
\\ \midrule
Full-ft & $86.6$ & $87.3$ & $88.2$ & $88.4$ & $91.2$ \\
\our\ (4-bits) & $83.2$ & $85.9$ & $87.4$ & $87.8$ & $90.7$ \\
\it Diff & \it -3.4 & \it -1.4 & \it -0.8 & \it -0.6 & \it -0.5 \\
\bottomrule
\end{tabular}}
\end{center}
\caption{Results in low-resource scenario. We randomly sample subsets from MNLI as training set and test on the standard validation set. }
\label{tab:data}
\end{table}

%% file: tables/baseline.tex
\begin{table*}[t]
\begin{center}
\setlength{\tabcolsep}{1.2mm}{
\scalebox{0.9}{\begin{tabular}{llllllllllll}
\toprule
\multirow{2}{*}{ Methods} & \multirow{2}{*}{ Bits}  &  Trainable &\multirow{2}{*}{ CoLA} &\multirow{2}{*}{ MNLI} &\multirow{2}{*}{ MRPC} &\multirow{2}{*}{ QNLI} &\multirow{2}{*}{ QQP} &\multirow{2}{*}{ RTE} &\multirow{2}{*}{ SST-2} &\multirow{2}{*}{ STS-B} &\multirow{2}{*}{ Avg.} \\
&&  Params
\\ \midrule 
FT & $32$ & $302$M & $68.0$ & $90.2$ & $90.9$ & $94.7$ & $92.2$ & $86.6$ & $96.4$ & $92.4$ & $88.9$ \\
\cline{1-12}
QAT-vanilla & $4$ & $302$M & $66.8$ & $89.2$ & $89.0$ & $83.5$ & $91.1$ & $86.4$ & $95.6$ & $91.0$ & $86.6$\\
Qadapter & $8$ & $0.29$M & $55.4$ & $87.8$ & $86.7$ & $91.9$ & $90.5$ & $84.4$ & $93.6$ & $90.7$ & $85.1$ \\
AlphaTuning & $4$ & $1.18$M & $57.8$ & $88.7$ & $88.6$ & $93.2$ & $91.2$ & $84.8$ & $95.2$ & $91.2$ & $86.3$\\
PreQuant & $4$ & $1.47$M & $67.3$ & $89.4$ & $89.0$ & $93.2$ & $91.1$ & $84.7$ & $95.4$ & $90.8$ & $87.6$\\
\bottomrule
\end{tabular}}}
\end{center}
\caption{Full results on the GLUE benchmark of the vanilla QAT and parameter-efficient quantization methods on $\textbf{RoBERTa}_{\text{large}}$.}
\label{tab:bas}
\end{table*}